\documentclass[submission]{IEEEtran}

\usepackage{graphicx}  

\usepackage{amsmath}   

\usepackage{multirow}
\usepackage{float,url}
\usepackage[left=0.71in,top=0.94in,right=0.71in,bottom=1.18in]{geometry}
\def\bf#1{\textbf{#1}}

\usepackage{fancyhdr}
\usepackage{listings}
\usepackage{color}
\usepackage{epsfig}
\usepackage{algorithmic,algorithm,cite,endnotes,subfigure,balance,amssymb}
\usepackage{multicol}

\newtheorem{lemma}{Lemma}[section]
\newtheorem{proposition}{Proposition}[section]
\floatstyle{ruled}
\newfloat{algo}{thp}{lop}
\floatname{algo}{Algorithm}

\begin{document}
\title{Fast Optimal Joint Tracking-Registration for Multi-Sensor Systems}
\author{Shuqing~Zeng,~\IEEEmembership{Member}
\thanks{Manuscript was drafted in May 28, 2010, and published in May 5, 2011. 
}
}
%
%
%
\markboth{Extended version of IEEE T. Instrumentation and Measurement 60(10): 3461-3470 (2011)}{S. Zeng}
%




\maketitle

\begin{abstract}
Sensor fusion of multiple sources plays an important role in vehicular systems to achieve refined target position and velocity estimates. In this article, we address the general registration problem, which is a key module for a fusion system to accurately correct systematic errors of sensors. A fast maximum a posteriori (FMAP) algorithm for joint registration-tracking (JRT) is presented. The algorithm uses a recursive two-step optimization that involves orthogonal factorization to ensure numerically stability. Statistical efficiency analysis based on Cram\`{e}r-Rao lower bound theory is presented to show asymptotical optimality of FMAP. Also, Givens rotation is used to derive a fast implementation with complexity $O(n)$ with $n$ the number of tracked targets. Simulations and experiments are presented to demonstrate the promise and effectiveness of FMAP.
\end{abstract}

\section{INTRODUCTION}
Recently, active safety driver assistance (ASDA) systems such as adaptive cruise control (ACC) and pre-crash sensing (PCS) systems \cite{Bishop2005} have drawn considerable attention in intelligent transportation systems (ITS) community. To obtain the necessary information of surround vehicles for ASDA systems, multiple sensors including active radars, lidars, and passive cameras are mounted on vehicles. Those sensor systems in a vehicular system are typically calibrated manually. However, sensor orientation and signal output may drift during the life of the sensor, such that the orientation of the sensor relative to the vehicular frame is changed. When the sensor orientation drifts, measurements become skewed relative to the vehicle \cite{RLuo2002}. When there are multiple sensors, this concern is further complicated. It is thus desirable to have sensor systems that automatically align sensor output to the vehicular frame \cite{Alempijevic2007,Ashokaraj2009}.

Two categories of approaches have been attempted to address the registration problem. The first category decouples tracking and registration into separate problems. In \cite{Zhou1997,Cruz1992,Sviestins2002,XLin2005,Jeong2008}, filters are designed to estimate the sensor biases by minimizing the discrepancy between measurements and associated fused target estimates, from a separated tracking module. However, these methods are not optimal in term of Cramer-Rao bounds~\cite{Tichavsky1998}. In the second category, the approaches jointly solve for target tracking and sensor registration. For example, \cite{Okello1996,Ong2002,Okello2003,Okello2004,Vermaak2005} have applied extended Kalman filtering (EKF) to an augmented state vector combining the target variables and sensor system errors in aerospace applications. In \cite{Challa2002} a fixed-lag smoothing framework is employed to estimate the augmented state vector to deal with communication jitters in networked sensors. Unscented Kalman filter (UKF) in \cite{Leung2004} and expectation-maximization based interacting multiple model (IMM) in \cite{Huang2005} have been applied to the augmented state vector in vehicle-to-vehicle cooperative driving systems. In \cite{Rigatos2009} the particle filter outperforms EKF at the cost of more demanding computations in state estimation for industrial systems. However, there are a few difficulties with these approaches. Since the registration parameters are constant, the error model of state-space is degenerated. This not only makes the estimation problem larger-leading to higher computational cost (complexity $O(n^3)$ with $n$ denoting number of targets), but also results degenerated covariance matrices for the process noise vectors due to numerical instability inherent to EKF, UKF, and the variants.

In this article, we propose a fast maximum a posteriori (FMAP) registration algorithm to tackle the problems. We combine all the measurement equations and process equations to form a linearized state-space model. The registration and target track estimates are obtained by maximizing a posterior function in the state space. The performance of FMAP estimates is examined using Cram\`{e}r-Rao lower bound theory. By exploiting the sparsity of the Cholesky factor of information matrix, a fast implementation whose complexity scales linearly with the numbers  of targets and measurements is derived.

The rest of this article is organized as follows. Section \ref{SC:3} is devoted to the algorithm derivation. An illustrative example is given in Section \ref{SC:5}. The results of simulation and experiment are presented in Sections \ref{SC:6} and \ref{SC:Experiment}, respectively. Finally we give concluding remarks in Section \ref{SC:7}.


\begin{figure}[b]
    \centering{
    \includegraphics[width=4cm]{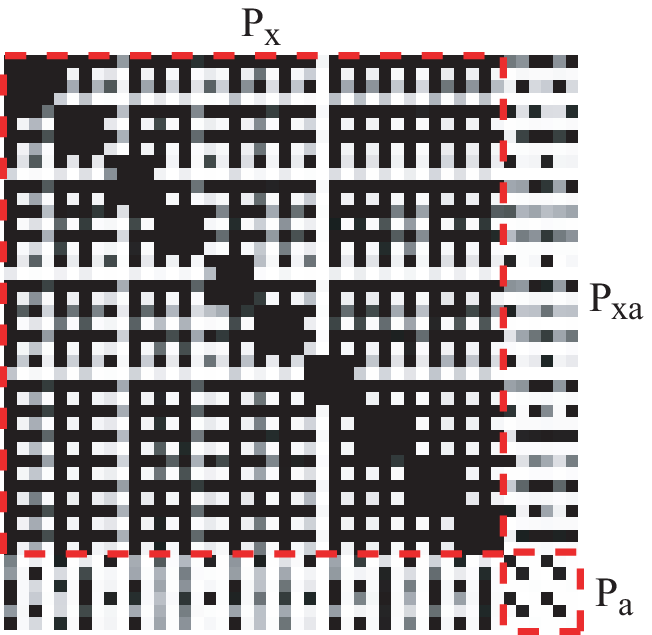}
    \includegraphics[width=4cm]{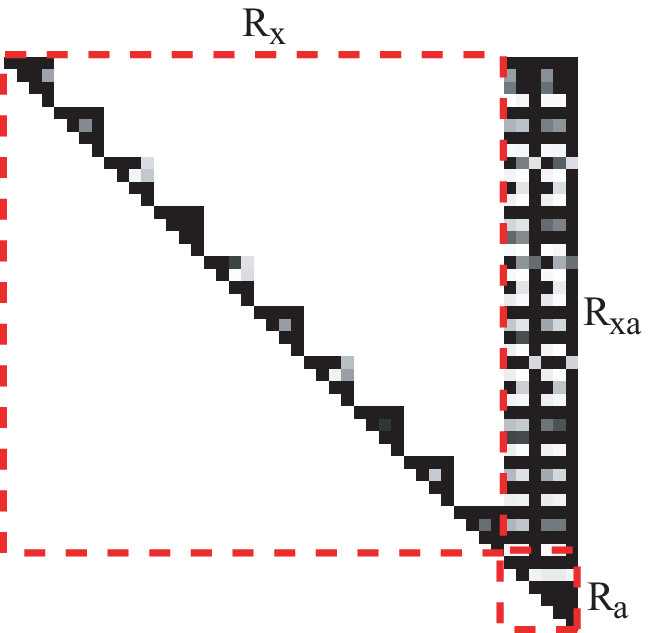}}\\
    \centering{\makebox[4cm]{(a)} \makebox[4cm]{(b)}}
    \caption{\protect\small  Typical snapshot of correlation matrices for JTR with 10 tracks $x$ and six registration parameters $a$. (a) a correlation matrix $\hat{P}$ of EKF (normalized). (b) Normalized Cholesky factor $\hat{R}$ of inverse covariance}
    \label{FG:correlation}
\end{figure}

\section{Algorithm Derivation} \label{SC:3}
In this section, we address the computational issues in using an EKF or UKF to solve joint tracking-registration (JTR). Fig. \ref{FG:correlation} shows the results of a joint problem with ten tracks and six registration parameters. The normalized covariance of the joint state from EKF is visualized in Fig. \ref{FG:correlation}(a). Dark entries indicate strong correlations. It is clear that not only the tracks $x$ and the registration $a$ are correlated but also each pair of tracks in $x$ is mutually correlated. The checkerboard appearance of the joint covariance matrix reveals this fact. Therefore, the approximation that ignores the off-diagonal correlated entries \cite{Zhou1997,Sviestins2002} is not asymptotical optimal.

A key insight that motivates the proposed approach is shown in Fig. \ref{FG:correlation}(b). Shown there is the Cholesky factor\footnote{The Cholesky factor $R$ of a matrix $P$ is defined as $P=R^tR$. A semi-positive definite matrix can be decomposed into its Cholesky factor.} of the inverse covariance matrix (also known as \emph{information matrix}) normalized like the correlation matrix. Entries in this matrix can be regarded as constraints, or connections, between the locations of targets and registration parameters. The darker an entry is in the display, the stronger the connection is. As this illustration suggests, the Cholesky factor $R$ not only appears sparse but also is nicely structured. The matrix is only dominated by the entries within a track, or the entries between a track and the registration parameters. The proposed FMAP algorithm exploits and maintains this structure throughout the calculation. In addition, storing a sparse factor matrix requires linear space. More importantly, updates can be performed in linear time with regard to the number of tracks in the system.

The sparsity of information matrix has been widely used to derive fast implementations of robotic Simultaneous Localization and Mapping (SLAM) \cite{Lu1997,Bailey2006,Kaess2007,LiuThrun2003,Frese2005}. In those approaches the authors insightfully observed that the resulting information matrix is sparse if the measurements involve only ``local" variables.
However the sparsity is destroyed in time-propagation steps where old robotic poses are removed from the state representation by marginalization. Approximations (e.g., \cite{LiuThrun2003,Frese2001}) are needed to enforce sparsity during the marginalization.

Although inspired by SLAM information filters, FMAP is different in the following aspects:
\begin{itemize}
\item JTR and SLAM are different problems despite the similarity between their system dynamics equations. The number of stationary landmarks dominates the time complexity in the SLAM case. On the other hand, the number of tracked targets that are in stochastic motion determines the time complexity in the JTR case.
\item Unlike SLAM, FMAP needs no approximation to enforce the sparsity in the measurement update and time propagation, as illustrated in Fig. \ref{FG:correlation} (b).
\item FMAP recursively computes Cholesky factor of information matrix $R = \left[\begin{array}{cc} R_x & R_{xa} \\ 0 & R_a\end{array}\right]$ as shown in Fig. \ref{FG:correlation} (b), contrasting the fact that information matrix is computed in the SLAM case. In addition, marginalization of old tracked targets $x'$ does not destroy the sparsity of $R$.
\end{itemize}

Throughout this article italic upper and lower case letters are used to denote matrices and vectors, respectively. A Gaussian distribution is denoted by information array \cite{Bierman1977}. For example, a multivariate ${x}$ with density function ${N}(\bar{{x}},{Q})$ is denoted as $p(x)\propto e^{(-\frac{\|Rx-z\|^2}{2})}$ or the information array $[{R},{z}]$ in short, where ${z}=R\bar{x}$ and $Q=R^{-1}R^{-T}$.

\subsection{Joint State Space}
The setting for the JTR problem is that a vehicle, equipped with multiple sensors with unknown or partially unknown registration, moves through an environment containing a population of objects. The sensors can take measurements of the relative position and velocity between any individual object and the vehicle.

The objective here is to derive a joint dynamics model where \textit{n} tracks and registration parameters from \textit{k} sensors are stacked into one large state vector as below:
\[\label{GrindEQ__2_1_}
s=\left[\begin{array}{cccc} x_{1}^{T}  & {x_{2}^{T} } & {....} & {\begin{array}{cccc} {x_{n}^{T} } & {a_{1}^{T} } & {....} & {a_{k}^{T} } \end{array}} \end{array}\right]^{T}
\]
where the \textit{i}-th target track $x_{i},(i=1,\dots n)$ comprises a set of parameters, e.g., position, velocity, and acceleration, and the registration parameters for \textit{j}-th sensor $a_{j}, (j=1,\dots ,k)$ comprises of location error, an azimuth alignment error, and range offset.

The system dynamics equation for the state is expressed as:
\begin{equation} \label{GrindEQ__1_}
s(t+1)=f (s(t),w(t))
\end{equation}
where the function relates the state at time t to the state at time t+1; and where terms $w$ are vectors of zero-mean noise random variables that are assumed to have nonsingular covariance matrices.

The measurement process can be modeled symbolically as a function of target track ($x$), and registration parameter ($a$), such as
\begin{equation} \label{EQ:obsr}
o(t) = h(x(t),a(t)) + v(t)
\end{equation}
where $o(t)$ and $v(t)$ denote the measurements and the additive noise vectors at time instant $t$, respectively.

\subsection{Measurement update} \label{SC:meas_update}

The FMAP algorithm works with the posterior density function $p(s(t)\mid o_{(0:t)})$, where $o_{(0:t)}$ denotes a series of measurements $\{o(0),...,o(t)\}$ from the sensors.

Using the Bayes rule, we obtain the posterior function as
\begin{equation}\begin{split}
&p(s(t)\mid o_{(0:t)}) = p(s(t)\mid o_{(0:t-1)}, o(t)) \\
&=c_1(t) p(o(t)\mid o_{(0:t-1)},s(t)) p(s(t)  \mid o_{(0:t-1)})
\end{split}
\label{EQ:posterior2}
\end{equation}
with $c_1(t)$ denotes the normalization factor. 
Typically, we will assume that measurements at time $t$ depend only on the current state $s(t)$\footnote{In Bayesian filtering, the state variables are designed such that they contain all information gathered from the past measurements. Thus $s(t)$ is a sufficient statistics of the past measurements $o_{(0:t-1)}$.}, and
\eqref{EQ:posterior2} can be written as
\begin{equation}
p(s(t)\mid o_{(0:t)})= c_1(t) p(o(t)\mid s(t)) p(s(t) \mid o_{(0:t-1)})
\label{EQ:posterior}
\end{equation}

Assuming the density functions are normally distributed, the prior density function\footnote{Unless it is necessary, we will not include time such as (t) in all the following equations.} $p(s \mid o_{(0:t-1)})$ can be expressed by information array $[\tilde{R},\tilde{z}]$, i.e.,
\begin{equation}{\small\begin{split}
&p(s \mid o_{(0:t-1)}) = \frac{|\tilde{R}|}{(2\pi)^{N_s/2}} \exp({-\frac{\left\|\tilde{R}s-\tilde{z}\right\|^2}{2}})\\
&= \frac{|\tilde{R}|}{(2\pi)^{N_s/2}} \exp({-\frac{\left\| \left[\begin{array}{cc} {\tilde{R}_{x} } & {\tilde{R}_{xa} } \\ {0} & {\tilde{R}_{a} } \end{array}\right]\left[\begin{array}{l} {x} \\ {a} \end{array}\right]-\left[\begin{array}{c} {\tilde{z}_{x} } \\ {\tilde{z}_{a} } \end{array}\right]\right\| ^{2}}{2}})
\end{split}}
\label{EQ:prior_pdf}
\end{equation}
where $|\tilde{R}|$ is the determinant of $\tilde{R}$, and $N_s$ is the dimension of $s$.

Linearizing  \eqref{EQ:obsr} using Taylor expansion in the neighborhood $[x^{*} ,a^{*} ]$, produces:
\begin{equation} \label{ZEqnNum185990}
o=C_{x} x+C_{a} a+u_1+v
\end{equation}
with $u_1=h^{*} -C_{x} x^{*} -C_{a} a^{*} $, $h^{*} =h(x^{*} ,a^{*} )$, and Jacobian matrices $C_{x} $ and $C_{a} $.  Without loss of generality, the covariance matrix of $v$ is assumed to be an identity matrix\footnote{If not, the noise term $v$ in  \eqref{ZEqnNum185990} can be transformed to a random vector with identity covariance matrix.  Let $\mbox{cov}\{ v\} =R_{v} $ denote the covariance matrix of the measurement model.  Multiplying both sides of  \eqref{ZEqnNum185990} by $L_{v} $, the square root information matrix of $R_{v} $, results in a measurement equation with an identity covariance matrix.}. Thus, the measurement density function can be written as
\begin{equation}
{\small \begin{split}
&p(o\mid s) = \frac{1}{(2\pi)^{N_o/2}} \\ &\cdot \exp({-\frac{ \left\| \left[ \begin{array}{cc} C_{x} & C_{a}\end{array}\right] \left[\begin{array}{c} x\\a \end{array}\right]-\begin{array}{c}(o-u_1)\end{array}\right\| ^{2}}{2}})\end{split}}\label{EQ:meas_pdf}\end{equation}
with $N_o$ being the dimension of $o$.

The negative logarithm of \eqref{EQ:posterior} is given by
\begin{equation}\begin{split}
J_t &= -\log p(s \mid o_{(0:t)})
\\&= -\log p(o\mid s) -\log p(s\mid o_{(0:t-1)}) -\log (c_1)
\end{split}
\label{EQ:neglog}
\end{equation}
Plugging in Eqs. \eqref{EQ:prior_pdf} and \eqref{EQ:meas_pdf}, \eqref{EQ:neglog} becomes
\begin{equation} \label{ZEqnNum288626}
J_{t} = \frac{\left\| \left[\begin{array}{cc} {\tilde{R}_{x} } & {\tilde{R}_{xa} } \\ {0} & {\tilde{R}_{a} } \\ {C_{x} } & {C_{a} } \end{array}\right]\left[\begin{array}{c} {x} \\ {a} \end{array}\right]-\left[\begin{array}{c} {\tilde{z}_{x} } \\ {\tilde{z}_{a} } \\ {o-u_1} \end{array}\right]\right\| ^{2}}{2}
+ c_2
\end{equation}
with $c_2$ denoting terms not depending on $(x,a)$.

The principle of the maximum a posteriori (MAP) estimation is to maximize the posterior function with respect to the unknown variables $(x,a)$. Clearly, the maximization process is equivalent to minimization of the squared norm in  \eqref{ZEqnNum288626}.
Ignoring the constant term, the right hand side (RHS) of \eqref{ZEqnNum288626} can be written as a matrix $X$ expressed as:
\begin{equation}
X=\left[\begin{array}{ccc} {\tilde{R}_{x} } & {\tilde{R}_{xa} } & {\tilde{z}_{x} } \\ {0} & {\tilde{R}_{a} } & {\tilde{z}_{a} } \\ {C_{x} } & {C_{a} } & {o-u_1} \end{array}\right]
\label{EQ:X-matrix}
\end{equation}
\textit{X} can be turned into an upper triangular matrix by applying an orthogonal transformation $\hat{T}$:
\begin{equation} \label{ZEqnNum330746}
\hat{T}X=\left[\begin{array}{ccc} {\hat{R}_{x} } & {\hat{R}_{xa} } & {\hat{z}_{x} } \\ {0} & {\hat{R}_{a} } & {\hat{z}_{a} } \\ {0} & {0} & {e } \end{array}\right]
\end{equation}
where $e$ is the residual that reflects the discrepancy between the model and measurement. Applying orthogonal $\hat{T}$ to the quadratic term in \eqref{ZEqnNum288626}, produces\footnote{The least-squares $\|Rs-z\|^2$ is invariant under an orthogonal transformation $T$, i.e., $\|T(Rs-z)\|^2 = (Rs-z)^tT^tT(Rs-z) = (Rs-z)^t(Rs-z) = \|Rs-z\|^2$.}:
\begin{equation} \label{ZEqnNum432510}
{\small
 \begin{array}{l} J_{t} =\frac{\left\| \left[\begin{array}{cc} {\hat{R}_{x} } & {\hat{R}_{xa} } \\ {0} & {\hat{R}_{a} } \end{array}\right]\left[\begin{array}{c} {x} \\ {a} \end{array}\right]-\left[\begin{array}{c} {\hat{z}_{x} } \\ {\hat{z}_{a} } \end{array}\right]\right\| ^{2}}{2} + c_3
 \end{array}
 }
\end{equation}
with $c_3=c_2+\frac{1}{2}\|e\|^2$ denoting the constant term with respect to variables $(x,a)$.

Because $J_t$ is the negative logarithm of $p(s \mid o_{(0:t)})$, the posterior density can be written as
\begin{equation}
{\small
p(s \mid o_{(0:t)}) \propto e^{-\frac{\left\| \left[\begin{array}{cc} {\hat{R}_{x} } & {\hat{R}_{xa} } \\ {0} & {\hat{R}_{a} } \end{array}\right]\left[\begin{array}{c} {x} \\ {a} \end{array}\right]-\left[\begin{array}{c} {\hat{z}_{x} } \\ {\hat{z}_{a} } \end{array}\right]\right\| ^{2}}{2}}}
\label{EQ:posterior_pdf}\end{equation}
or in the information-array form
\begin{equation} \label{ZEqnNum934500}
\left[\hat{R},\hat{z}\right]=\left[\begin{array}{ccc} {\hat{R}_{x} } & {\hat{R}_{xa} } & {\hat{z}_{x} } \\ {0} & {\hat{R}_{a} } & {\hat{z}_{a} } \end{array}\right]
\end{equation}

Since the estimate of the state variable $s$ can be written as $\hat{s}=\hat{R}^{-1}\hat{z}$, \eqref{ZEqnNum934500} allows us to solve the estimates of the track variables $\hat{x}$ and registration parameters $\hat{a}$ by back-substitution using $\hat{R}$ and right hand side $\hat{z}$, i.e.,
\begin{equation}
\left[\begin{array}{cc} {\hat{R}_{x} } & {\hat{R}_{xa} } \\ {0} & {\hat{R}_{a} } \end{array}\right]\left[\begin{array}{c} x \\ a\end{array} \right] = \left[\begin{array}{c} \hat{z}_x \\ \hat{z}_a \end{array}\right] \label{EQ:back_substitution}
\end{equation}

\subsection{Time propagation} \label{SC:time_propagation}
The prior density $p(s(t+1)\mid o_{(0:t)})$ at time $t+1$ can be inferred from the system dynamics and posterior function $p(s\mid o_{(0:t)})$ at time $t$. Assuming registration is time-invariant the linear approximation of the system dynamics in  \eqref{GrindEQ__1_} in the neighborhood $[s^{*} ,w^{*} ]$ can be expressed as,
\begin{equation} \label{GrindEQ__2_14_}
{\scriptsize
\begin{array}{l}
\left[\begin{array}{c}x(t+1)\\a(t+1)\end{array}\right]=\left[\begin{array}{cc}\Phi_x& 0\\0&I\\\end{array}\right]\left[\begin{array}{c} x\\a\end{array} \right]
+ \left[\begin{array}{c} G_x \\0 \end{array}\right] w+ \left[\begin{array}{c} u_2\\0\end{array} \right]
\end{array}}
\end{equation}
where $\Phi_x$ and $G_x$ are Jacobian matrices; and the nonlinear term $u_2 =f (s^{*} ,w^{*} )-\Phi_x x^{*} -G_x w^{*}$.

If variables $s$ and $w$ are denoted by the information arrays in  \eqref{ZEqnNum934500} and $[R_{w},z_{w}]$, respectively, the joint density function given the measurements $o_{(0:t)}$  can be expressed in terms of $x(t+1)$ and $a(t+1)$ as\footnote{Detailed derivation is shown in Lemma \ref{lemma_NAIK2} in Appendix A.}
\begin{equation}
{\tiny p(s(t+1),w\mid o_{(0:t)})\propto e^{-\frac{\left\| A s_w - b\right\| ^{2} }{2}}\label{ZEqnNum434705}}
\end{equation}
with {\small $A=\left[\begin{array}{ccc} {R_{w}} & {0} & {0} \\ {-\hat{R}_{x} \Phi _{x}^{-1} G_{x} } & {\hat{R}_{x} \Phi _{x}^{-1} } & {\hat{R}_{xa}} \\ {0} & {0} & {\hat{R}_{a} } \end{array}\right]$}, {\small $s_w=\left[\begin{array}{c} {w} \\ {x(t+1)} \\ {a(t+1)} \end{array}\right]$}, and {\small$b=\left[\begin{array}{c} {z_{w} } \\ {\hat{z}_{x} +\hat{R}_{x} \Phi _{x}^{-1} u_2} \\ {\hat{z}_{a}} \end{array}\right]$}.

The quadratic exponential term in  \eqref{ZEqnNum434705} can be denoted as matrix \textit{Y}, expressed as
\begin{equation}
Y =\left[\begin{array}{cc}A& b\end{array}\right]
\label{EQ:Y-matrix}
\end{equation}
The matrix \textit{Y} can be turned into a triangular matrix through an orthogonal transformation $\tilde{T}$, i.e.,
\begin{equation}
\tilde{T}Y= \left[\begin{array}{cc}\tilde{A}& \tilde{b}\end{array}\right]\label{ZEqnNum354069}
\end{equation}
with {\small $\tilde{A} = \left[\begin{array}{cccc} {\tilde{R}_{w} (t+1)} & {\tilde{R}_{wx} (t+1)} & {\tilde{R}_{wa} (t+1)} \\ {0} & {\tilde{R}_{x} (t+1)} & {\tilde{R}_{xa} (t+1)}\\ {0} & {0} & {\tilde{R}_{a} (t+1)}  \end{array}\right]$} and {\small $\tilde{b} = \left[\begin{array}{c}
{\tilde{z}_{w} (t+1)}\\
{\tilde{z}_{x} (t+1)} \\
{\tilde{z}_{a} (t+1)} \end{array}\right]$}.

Given the measurements $o_{(0:t)}$, the prior function $p(s(t+1)\mid o_{(0:t)})$ can be produced by marginalization on variable $w$. Applying Lemma~\ref{lemma_marginal} in Appendix A, we obtain
\begin{equation}
{\scriptsize \begin{split}
&p(s(t+1)\mid o_{(0:t)}) =c_4 \exp{-\frac{\left\|\tilde{R}(t+1)s(t+1)-\tilde{z}(t+1)\right\|^2}{2}} \\&=c_4 e^{-\frac{\left\|{\left[\begin{array}{cc} {\tilde{R}_{x} (t+1)} & {\tilde{R}_{xa} (t+1)}  \\ {0} & {\tilde{R}_{a} (t+1)} \end{array}\right]\left[\begin{array}{c} x(t+1)\\a(t+1) \end{array}\right]-\left[\begin{array}{c}{\tilde{z}_{x} (t+1)}\\{\tilde{z}_{a} (t+1)}\end{array}\right]}\right\|^2}{2}}\end{split}}
\nonumber
\end{equation}
with $c_4$ being the normalization factor.

Therefore, we obtain
the updated prior information array $\left[\tilde{R}(t+1),\tilde{z}(t+1)\right]$at time $t+1$ as:
 \begin{equation}\left[\begin{array}{ccc} {\tilde{R}_{x} (t+1)} & {\tilde{R}_{xa} (t+1)} & {\tilde{z}_{x} (t+1)} \\ {0} & {\tilde{R}_{a} (t+1)} & {\tilde{z}_{a} (t+1)} \end{array}\right] \label{2.20}
 \end{equation}
The derivation leading to \eqref{2.20} illustrates the purposes of using the information array $[\tilde{R},\tilde{z}]$, which is recursively updated to $[\tilde{R}(t+1),\tilde{z}(t+1)]$ at time instant $t+1$.

One can verify (c.f., \cite[VI.3]{Bierman1977}) that the covariance matrix $\tilde{P}(t+1) =\Phi \hat{P} \Phi^T + G Q G^T$ where $\tilde{P}(t+1) = \tilde{R}^{-1}(t+1)\tilde{R}^{-T}(t+1)$, $\hat{P}=\hat{R}^{-1}\hat{R}^{-T}$, $Q=R_w^{-1}R_w^{-T}$, $G=\left[\begin{array}{cc} G_x^T & 0 \end{array}\right]^T$ and $\Phi=\left[\begin{array}{cc}\Phi_x& 0\\0&I\\\end{array}\right]$. In addition, $\tilde{s}(t+1) = \Phi \hat{s} + G \bar{w} + u$ where $\tilde{s}(t+1)=\tilde{R}^{-1}(t+1)\tilde{z}(t+1)$, $\bar{w}$ denotes the mean of $w$, and $u=\left[\begin{array}{cc} u_2^T & 0\end{array} \right]^T$.

\subsection{Algorithm} \label{SC:Algorithm}
\begin{figure}[t]
    \centering
    \includegraphics[width=7cm]{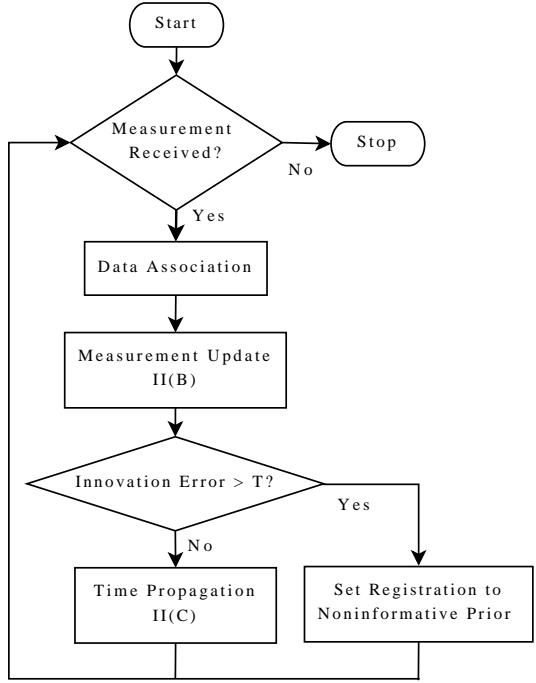}
    \caption{\protect\small Flow chart of the fast maximum a posteriori (FMAP) algorithm.}
    \label{FG:Architecture}
\end{figure}

Fig.~\ref{FG:Architecture} illustrates the flow chart of FMAP. The algorithm is started upon reception of sensor data, and the state variables of every track and registration parameters are initialized using zero-mean noninformative distributions, respectively. A \textit{data association} module matches the sensor data with the predicted location of targets. The \textit{measurement update} module combines the previous estimation (i.e., prior) and new data (i.e., matched measurement-track pairs), and updates target estimation and registration. FMAP checks whether the discrepancy between measurements and predictions (innovation error) is larger than a threshold $T$. A change of registration parameters is detected when the threshold is surpassed, and the prior of the sensor registration is reset to the noninformative distribution. This is typically occurred when the sensor's pose is significantly moved. The \textit{time propagation} module predicts the target and registration in the next time instant based on the dynamics model \eqref{GrindEQ__2_14_}.

The complete specification of the proposed algorithm is given in Algorithm 1. Note that the prior $[\tilde{R}_0, \tilde{z}_0]$ at time 0 is initialized as $\tilde{R}_0 = \varepsilon I$ and $\tilde{z}_0 = \mathbf{0}$ where $\varepsilon$ is a small positive number, and $I$ is an identity matrix of appropriate dimension.

Although static registration $a$ is assumed in the derivation (i.e., Eq.~\eqref{GrindEQ__2_14_}), the case in which $a$ in stochastic motion can be easily accommodated. As shown in Fig.~\ref{FG:dynamicregistr}, the step change of $a$ can be easily detected by applying threshold checking to innovation error curve. Once a change is detected, we set the registration prior to the noninformative distribution, i.e., $\tilde{z}_a=\mathbf{0}$ and $\tilde{R}_a = \varepsilon I$. This forces FMAP to forget all past information regarding registration and to trigger a new estimation for $a$. Since $a$ rarely occurs and sufficient duration exists between two changes, as shown in Fig.~\ref{FG:dynamicregistr}, we can approximate the dynamics of sensor by a piecewise static time propagation model.

\begin{algo}
\parbox{3.2in}{
\noindent{\bf Algorithm 1:} FMAP update\\
{\small
\begin{algorithmic}[1]
\REQUIRE Given prior at instant $t$ (i.e., previous results and its uncertainty measure) expressed as information array $[\tilde{R}, \tilde{z}]$ and measurements $o$; the system dynamical equation \eqref{GrindEQ__1_} and measurement equation \eqref{EQ:obsr}.
\ENSURE The updated estimate of $s$, expressed by $\hat{s}$
\medskip
\STATE Compute $C_x$, $C_a$, and $u_1$.
\STATE Plug prior $[\tilde{R}, \tilde{z}]$; sensor measurement matrices $C_x$ and $C_a$; and vectors $u_1$ and $o$ into matrix $X$ (c.f.,  \eqref{EQ:X-matrix}).
\STATE Factorize $X$ (c.f., Algorithm 2).
\STATE Derive the posterior density information array $[\hat{R},\hat{z}]$ as shown in \eqref{ZEqnNum934500} (Measurement update).
\STATE Compute the update of tracking and registration as $\hat{s}=\hat{R}^{-1}\hat{z}$ (c.f.,\eqref{EQ:back_substitution}).
\STATE Compute $\Phi_x$, $G_x$, and $u_2$.
\STATE Plug the posterior information array $[\hat{R}, \hat{z}]$, $R_w$, $\Phi_x$, and $G_x$ into $Y$.
\STATE Factorize $Y$ (c.f., Algorithm 3).
\STATE Derive prior information array $[\tilde{R}(t+1), \tilde{z}(t+1)]$ for time $t+1$ (c.f., \eqref{2.20}), which can be utilized when the new sensor measurements are available (Time propagation).
\end{algorithmic}
}}
\end{algo}

\subsection{Statistical Efficiency Analysis}
The Cram\`{e}r-Rao lower bound (CRLB) is a measure of statistical efficiency of an estimator. Let $p(o_{(0:t)},s)$ be the joint probability density of the state (parameters) $s$ at time instant $t$ and the measured data $o_{(0:t)}$, and let $g(o_{(0:t)})$ be a function of an estimate of $s$. The CRLB for the estimation error has the form
\begin{equation}
P \equiv \mbox{E}\{[g(o_{(0:t)})-s][g(o_{(0:t)})-s]^t\} \geq J_t \label{EQ:CRLB}
\end{equation}
where $J_t$ is the Fisher information matrix with the elements
\[
J_t^{(ij)} = \mbox{E}\left\{\frac{\partial^2 \log p(o_{(0:t)},s) }{\partial s_i \partial s_j}\right\}
\]

\begin{proposition}
Considering the system defined by \eqref{GrindEQ__1_} and \eqref{EQ:obsr}, the Fisher information matrix of the system $J_t$ at time $t$ is identical to $[\hat{R}(t)]^t[\hat{R}]$, with $\hat{R}(t)$ defined in \eqref{ZEqnNum934500}.
\end{proposition}
\begin{proof}
The joint probability density can be derived from the equality $p(o_{(0:t)},s) = p(s|o_{(0:t)})p(o_{(0:t)})$. Since $p(o_{(0:t)})$ is a function of measured data, not depending on the state $s$; therefore, we have $J_t = \mbox{E}\{\Delta \log p(s|o_{(0:t)}\}$
where $\Delta$ denotes the second-order partial derivative, i.e., $\Delta L(s) = \frac{\partial L(s)}{\partial s} \frac{\partial L(s)}{\partial s}^t$.
Plugging \eqref{EQ:posterior_pdf} in, the logarithm of the posterior function $\log p(s|o_{(0:t)})$ reads
\[
-\log p(s|o_{(0:t)})  = c_0 + \frac{\left\|\hat{R}s-\hat{z} \right\| ^{2}}{2}
\]
where $c_0$ denotes a constant independent of $s$. Then the Fisher information matrix $J_t$ reads $J_t = [\hat{R}(t)]^t[\hat{R}]$.
\end{proof}

If $s$ is estimated by $g(s)= \mbox{E}(s|o_{(0:t)}) = \hat{R}^{-1} \hat{z}$, then \eqref{EQ:CRLB} is satisfied with equality. Therefore the FMAP algorithm is optimal in the sense of CRLB.

\subsection{Fast Implementation} \label{SC:4}
We have observed the FMAP algorithm comprises two factorization operations outlined in Eqs. \eqref{ZEqnNum330746} and \eqref{ZEqnNum354069}, and back substitution operation \eqref{EQ:back_substitution}.  However, directly applying matrix factorization techniques (e.g., QR decomposition) can be as computationally ineffective as the UKF since the complexity of QR is $O(n^{3})$ (\textit{n} denotes number of targets).

\begin{figure}[htbp]
    \centering
    \includegraphics[width=8.5cm]{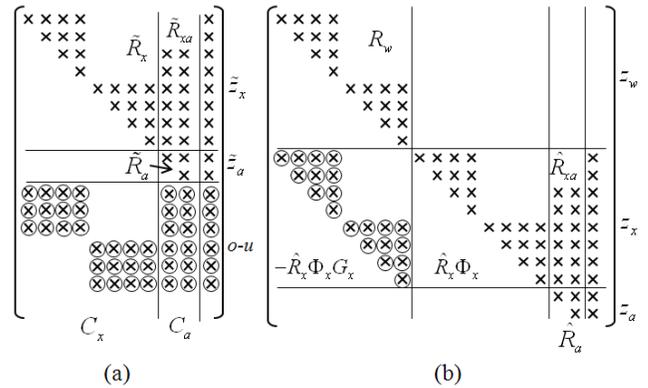}
    \caption{\protect\small  Triangular factorization (a) Example $X$ matrix (b) Example $Y$ matrix}
    \label{FG:meas_update}
\end{figure}

As shown in Fig.~\ref{FG:correlation}, we have noted that the matrices in \eqref{EQ:X-matrix} are nicely structured. Fig.~\ref{FG:meas_update}(a) illustrates an example schematically with two tracks, two registration parameters, and six measurements. The non-zero elements of the matrix in  \eqref{EQ:X-matrix} are denoted by crosses; and a blank position represents a zero element.

Givens rotation~\cite{Golub96} is used to eliminate the non-zero elements of the matrix $C_x$ in \eqref{EQ:X-matrix}, shown in Fig.~\ref{FG:meas_update}(a) as crosses surrounded by circles. Givens rotation is applied from the left to the right and for each column from the top to the bottom. Each non-zero low-triangular element in the \textit{i}-th row of $C_{x}$, is combined with the diagonal element in the same column in the matrix block $\tilde{R}_{x}$ to construct the rotation.  If the element in $C_x$ is zero, then no rotation is needed.

Consequently, the algorithm to factorize $X$ can be written as Algorithm 2. Note that Algorithm 2 is an in-place algorithm, which uses a small, constant amount of extra storage space.

\begin{algo}
\parbox{3.2in}{
\noindent{\bf Algorithm 2:} Factorization of $X$\\
{\small
\begin{algorithmic}[1]
\REQUIRE Given $X$ defined in \eqref{EQ:X-matrix}.
\ENSURE  Output an upper triangular matrix $X' = \hat{T}X$ such that $X'^tX' = X^tX$, with $\hat{T}$ an orthogonal matrix.
\medskip
\STATE Let $L$ be the dimension of $s$.
\FORALL {Column $j$ in $C_x$}
\FORALL {Row $i$ in $C_x$}
\IF {$C_x(i,j)\neq 0$}
    \STATE Let $\alpha=\tilde{R}_x(i,i)$ and $\beta=C_x(i,j)$.
    \STATE Construct the rotation $\mbox{Rot}(\alpha,\beta) = \left[\begin{array}{cc}\alpha/r&\beta/r\\-\beta/r&\alpha/r\end{array}\right]$ with $r=\sqrt{\alpha^2+\beta^2}$.
    \FORALL {$k$ such that  $j\leq k \leq N$, with $N$ denoting the last column of $X$}
    \IF {$X(i,k) \neq 0$ or $X(i+L,k) \neq 0$ }
    \STATE Let $d=\left[\begin{array}{c}X(i,k)\\X(i+L,k)\end{array}\right]$.
    \STATE Apply the rotation Rot$(\alpha,\beta)$ to $d$, i.e., $d' = \mbox{Rot}(\alpha,\beta)d$.
    \STATE Save $d'$ back to $X$ in the same location as $d$.
    \ENDIF
    \ENDFOR
\ENDIF
\ENDFOR
\ENDFOR
\STATE Apply QR decomposition to the sub-matrix {\small $\Lambda=\left[\begin{array}{cc}\tilde{R}_a & \tilde{z}_a\\C_a&o-u_1\end{array}\right]$}, i.e., $\Lambda = T_{qr}U$ with $T_{qr}$ and $U$ being the orthogonal and upper triangular matrices, respectively.
\STATE Write the result $U$ back to $X$ in the same location as $\Lambda$.
\end{algorithmic}
}}
\end{algo}

\begin{lemma} \label{lemma:2}
If $\tilde{R}_x$ in  \eqref{EQ:X-matrix} is a block-diagonal matrix, then 1) the result $\hat{R}_x$ in  \eqref{ZEqnNum330746} has the same form as $\tilde{R}_x$; and 2) the complexity of Algorithm 2 is $O(mk^2)$, where $m$ and $k$ denote the numbers of measurements and sensors, respectively.
\end{lemma}
\begin{proof}
The triangulation described in Algorithm 2 comprises of two steps: first turn $C_x$ into a zero matrix using Givens rotation; and second apply QR decomposition to the sub-matrix $\Lambda$. Note that $C_x$ in $X$ is sparse. This is seen by expanding the measurement equation \eqref{EQ:obsr} as
$o_{\kappa_{i} }^{(j)} =h(x_{i} ,a_{j} )+v_j$
where $o_{\kappa_{i} }^{(j)}$ denotes the measurements associated with the target track $x_{i}$ from the \textit{j}-th sensor; and $\kappa_{i}$ is the measurement association variable for the \textit{i}-th target track. Therefore, there is at most $O(N_{x} +kN_{a})$ nonzero entries in each paired set of rows of $\tilde{R}_x$ and $C_x$, where $N_{x}$ denotes the number of parameters for each tracked target; and $N_{a}$ denotes the number of registration parameters for each sensor. Since an off-diagonal zero entry $\varsigma$ in $\tilde{R}_x$ pairs up with a zero entry in $C_x$ in the rotation operations, the condition specified in Step 8 of Algorithm 2 is not satisfied, and then the off-diagonal entry $\varsigma$ remains zero after the rotation. Thus 1) is established.

We note that in each paired set of rows of $\tilde{R}_x$ and $C_x$, there are at most $O(N_x+kN_a)$ non-zero pairs. Therefore, in order to turn an element of $C_x$ to zero, at most $O(N_{x} +kN_{a} )$ rotations in Step 10 are needed. Providing that \textit{m} measurements are presented, $O(mN_{x})$ elements need to be eliminated to zero.  Givens rotation is applied a maximum $O(mN_{x}^{2} +mkN_{x} N_{a})$ times to transform $C_x$ into a zero matrix, which is equivalent to $O(mk)$ additive and multiplicative operations. In the second step, additional $O(mk^2)$ operations are needed. Therefore, the total complexity is $O(mk^2)$.
\end{proof}

Triangulation of  \eqref{ZEqnNum354069} can be treated similarly as that of  \eqref{ZEqnNum330746}. Fig.~\ref{FG:meas_update}(b) illustrates an example schematically with two tracks and two registration parameters.  The non-zero elements of the matrix \textit{Y} are denoted by crosses; and blank position represents a zero element. Givens rotation is used to eliminate the non-zero entries in the matrix $R_{xG}^d=-\hat{R}_x \Phi_x^{-1} G_x$, shown in Fig.~\ref{FG:meas_update}(b) as crosses annotated by circles.  Givens rotation is applied from the left to the right and for each column from the top to the bottom. Each non-zero entry in the \textit{i}-th row of the matrix $R_{xG}^d$ is paired with the element in the same column in $R_{w} $ to construct the rotation.

Similarly, the algorithm to factorize $Y$ is given as Algorithm 3.

\begin{algo}
\parbox{3.2in}{
\noindent{\bf Algorithm 3:} Factorization of $Y$\\
{\small
\begin{algorithmic}[1]
\REQUIRE Given $Y$ defined in \eqref{EQ:Y-matrix}.
\ENSURE  Output an upper triangular matrix $Y' = \tilde{T}Y$ such that $Y'^tY' = Y^tY$, with $\tilde{T}$ an orthogonal transformation.
\medskip
\STATE Let $R_{xG}^d = -\hat{R}_x \Phi_x^{-1} G_x$. Let $L$ be the number of rows in $R_w$.
\FORALL {Column $j$ in $R_{xG}^d$}
\FORALL {Row $i$ in $R_{xG}^d$}
\IF {$R_{xG}^d(i,j)\neq 0$}
    \STATE Let $\alpha=Y(i,i)$ and $\beta=R_{xG}^d(i,j)$.
    \STATE Construct the rotation $\mbox{Rot}(\alpha,\beta)$.
    \FORALL {$k$ such that  $j\leq k \leq N$, with $N$ denoting the last column of $Y$}
    \IF {$B(i,k) \neq 0$ or $B(i+L,k) \neq 0$ }
    \STATE Let $d=\left[\begin{array}{c}B(i,k)\\B(i+L,k)\end{array}\right]$
    \STATE Apply the rotation Rot$(\alpha,\beta)$ to $d$, i.e., $d' = \mbox{Rot}(\alpha,\beta)d$
    \STATE Save $d'$ back to $Y$ in the same location as $d$.
    \ENDIF
    \ENDFOR
\ENDIF
\ENDFOR
\ENDFOR
\STATE Let $R_x^d = \hat{R}_x \Phi_x^{-1}$. Let $M$ be the number of columns of $R_w$.
\FORALL {Column $j$ in $R_x^d$}
\FORALL {Row $i$ in $R_x^d$}
\IF {$R_x^d(i,j)\neq 0$}
    \STATE Let $\alpha=Y(i,i+M)$ and $\beta=R_{x}^d(i,j)$.
    \STATE Construct the rotation $\mbox{Rot}(\alpha,\beta)$.
    \FORALL {$k$ such that  $j+M  \leq k \leq N$, with $N$ denoting the last column of $Y$}
    \IF {$B(i,k) \neq 0$ or $B(i+L,k) \neq 0$ }
    \STATE Let $d=\left[\begin{array}{c}B(i,k)\\B(i+L,k)\end{array}\right]$
    \STATE Apply the rotation Rot$(\alpha,\beta)$ to $d$, i.e., $d' = \mbox{Rot}(\alpha,\beta)d$
    \STATE Save $d'$ back to $Y$ in the same location as $d$.
    \ENDIF
    \ENDFOR
\ENDIF
\ENDFOR
\ENDFOR
\end{algorithmic}
}}
\end{algo}

\begin{lemma} \label{lemma:3}
If $\hat{R}_x$ in \eqref{EQ:Y-matrix} is a block-diagonal matrix, then 1) the result $\tilde{R}_x(t+1)$ in  \eqref{ZEqnNum354069} has the same form as $\hat{R}_x$; and 2) the complexity of Algorithm 3 is $O(nk)$, with $n$ denoting the number of targets.
\end{lemma}
\begin{proof}
Each individual target has its own system dynamics equation. For the \textit{i}-th target, \eqref{GrindEQ__2_14_} can be expressed as:
\[x_{i} (t+1)=\Phi _{i} x(t)+G_{i} w_{i} (t)+u_{2i} \]
where $\Phi_{i} $and $G_{i} $ are Jacobian matrices; the nonlinear term $u_{2i} =f (x_{i}^{*} ,w_{i}^{*} )-\Phi _{i} x_{i}^{*} -G_{i} w_{i}^{*} $; and $w_{i}(t)$ is represented by the information array $[R_{wi},z_{wi}]$. Therefore, the corresponding collective quantities $\Phi_{x} $, $G_{x}$, and $R_w$ are in the same block-diagonal form. In addition, $\hat{R}_x\Phi_x^{-1}$ and $R_{xG}^d$ are in the same block-diagram form. Since each off-diagonal zero entry $\varsigma$ in $\hat{R}_x \Phi_x^{-1}$ pairs up with a zero entry in the rotation operations, the condition in Steps 8 and 24 is not satisfied for the pairs and, then, the off-diagonal entry $\varsigma$ remains zero. Thus 1) is established.

As Fig.~\ref{FG:meas_update}(b) shows, in the paired rows of $R_w$ and $R_{xG}^d$, there are at most $O(2N_{x} +kN_{a})$ non-zero pairs. Thus, to eliminate an element to zero, a maximum $O(2N_{x} +kN_{a})$ rotations in Steps 10 and 26 are needed. If \textit{n} tracked targets are considered, $O(nN_{x} ^{2})$ elements need to be eliminated to zero.  Givens rotation is applied a maximum $O(2nN_x^3+knN_aN_x^2)$ times to transform the matrix \textit{Y} into a triangular matrix, which is equivalent to $O(nk)$ additive and multiplicative operations. 2) is established.
\end{proof}

\begin{lemma}\label{lemma:4}
The complexity of the back substitution operation expressed in \eqref{EQ:back_substitution} is $O(n+k^3)$.
\end{lemma}
\begin{proof}
Since $\hat{R}_x$ is a block-diagonal matrix, \eqref{EQ:back_substitution} can be written as
\[
\left[\begin{array}{cccc} \hat{R}_{x_1} &\ldots &0 &\hat{R}_{x_1a}\\
 \vdots &\ddots & \vdots &\vdots \\
 0 &\ldots & \hat{R}_{x_n} & \hat{R}_{x_na} \\
 0 &\ldots & 0 & \hat{R}_a \end{array}\right]\left[\begin{array}{c}x_1\\\vdots\\x_n\\a \end{array}\right] =\left[\begin{array}{c} z_{x_1}\\\vdots\\z_{x_n}\\z_a\end{array}\right]
\]
where ${x_i}$ denotes the $i$-th target. One can verify that
\begin{equation}\begin{split}
a &= \hat{R}_a^{-1}\\
x_i &= \hat{R}_{x_i}^{-1}(\hat{z}_{x_i}-\hat{R}_{x_ia}a)
\end{split}
\label{EQ:backsbtk}
\end{equation}
$O((kN_{a})^3)$ and $O(nN_{x} ^{3} )$ operations are needed to solve $\hat{a}$ and $\hat{x_i}$ for $i=1,...,n$ in \eqref{EQ:backsbtk}, respectively. Let $\hat{P}_{x_i}$ and $\hat{P}_a$ be covariance matrices of $x_i$ and $a$. We can write
\begin{equation}\begin{split}
\hat{P}_a &= \hat{R}_a^{-1}\hat{R}_a^{-T} \\
\hat{P}_{x_i} &= \hat{R}_{x_i}^{-1}\hat{R}_{x_i}^{-T} + \hat{R}_a^{-1} \hat{R}_{x_ia} \hat{P}_a \hat{R}_{x_ia}^T \hat{R}_a^{-T}
\end{split}
\label{EQ:backsbtk_cov}
\end{equation}
Additional $O((kN_{a})^3)$ and $O(nN_{x} ^{3} )$ operations are needed to solve $\hat{P}_a$ and $\hat{P}_{x_i}$ for $i=1,...,n$ in \eqref{EQ:backsbtk_cov}, respectively.

Therefore, the total complexity is $O(n+k^3)$.
\end{proof}

To summarize, we establish the following proposition:
\begin{proposition}
The complexity of the FMAP algorithm is $O((m+n)k^2 + k^3)$, where $m$, $n$ and $k$ denote the numbers of measurement equations, targets and sensors.
\end{proposition}
\begin{proof}
The FMAP algorithm comprises of three matrix operations whose complexity are specified by Lemmas \ref{lemma:2}, \ref{lemma:3} and \ref{lemma:4}, respectively. Thus the total complexity is $O(mk^2+nk+n+k^3)=O((m+n)k^2 + k^3)$.
\end{proof}

Note that the number of sensors $k$ is a constant. The complexity in the above proposition is thus simplified as $O(n+m)$. This shows FMAP scales linearly with the number of measurements and with the number of target tracks.

\subsection{Handling Changes of the State Vector} \label{SC:change_dim}
So far, we have assumed the state vector $s=[x,a]$ has a fixed dimension. This means that we have the same number of targets in the whole observation span. However this is rarely true in practice. The dimension of $s$ changes due to new targets coming into or existing targets leaving the field-of-view of the sensors.

Once $s$ has been estimated at time instant $t$ (c.f., Section~\ref{SC:meas_update}), the prior distribution of $s(t+1)$ (inferred from $s$) can be easily obtained using the method described in Section~\ref{SC:time_propagation}. Therefore, the main problem is to compute the distribution of the updated state vector $s'(t+1)$ after changes of the state vector $s$.

In the following we define different vectors:
\begin{itemize}
\item $x^r$: all targets that are visible at instant $t$ and remain at instant $t+1$;
\item $x^c$: all new targets that are visible at instant $t+1$ but not visible at instant $t$;
\item $x^d$: all deleted targets that are visible at instant $t$ but not visible at instant $t+1$;
\item $s^r$: the concatenated vector of the remaining targets and registration parameters, i.e., $s^r=[x^r,a]$.
\end{itemize}
Note that the following relationships hold:
\[
 s(t+1) = \left[\begin{array}{c} x^d\\ s^r\end{array}\right], s'(t+1) = \left[\begin{array}{c} x^c\\ s^r \end{array}\right]
\]

Let $\Pi=\{\pi_1,...,\pi_d,\pi_{d+1},...,\pi_n\}$ be a permutation of $s(t+1)$ such that
\[s(t+1)=\left[\begin{array}{ccc}\overbrace{x_{\pi_1},...,x_{\pi_d}}^{x^d} & \overbrace{x_{\pi_{d+1}},...,x_{\pi_n},a}^{s^r} \end{array}\right]\]
So we can find a permutation matrix~\cite{Golub96} $P_\Pi=[P^{(1)}_\Pi, P^{(2)}_\Pi]$ such that
\[
    P_\Pi s(t+1) = \left[\begin{array}{c} P^{(1)}_\Pi s(t+1) \\ P^{(2)}_\Pi s(t+1) \\ a(t+1) \end{array} \right] = \left[ \begin{array} {c} x^d \\ x^r\\a(t+1)\end{array}\right]
\]

We can verify that the marginal distribution of $s^r$ as (c.f., Lemma~\ref{lemma_marginal} in Appendix A)
\[
s^r = \left[\begin{array}{c}x^r \\ a(t+1)\end{array}\right] \sim \left[\begin{array}{ccc} \tilde{R}^r_x & \tilde{R}^r_{xa} & \tilde{z}_x^r \\
                  0 & \tilde{R}_a(t+1) & \tilde{z}_a(t+1) \end{array} \right]
\]
with
\[
\begin{split}
\tilde{R}_x^r &= P^{(2)}_\Pi \tilde{R}_x(t+1)\left(P^{(2)}_\Pi\right)^T\\
\tilde{R}^r_{xa}&= P^{(2)}_\Pi \tilde{R}_{xa}(t+1)\\
\tilde{z}_{x}^r &= P^{(2)}_\Pi \tilde{z}_x (t+1) \\
\end{split}
\]
where $\tilde{R}_x(t+1)$, $\tilde{R}_{xa}(t+1)$, $\tilde{R}_a(t+1)$, $\tilde{z}_x (t+1)$, and $\tilde{z}_a(t+1)$ are defined in \eqref{2.20}. We note that $\tilde{R}_x^r$ is a block-diagonal matrix, i.e., $\tilde{R}_x^r = \mbox{diag}([\tilde{R}_{\Pi_{d+1}},...,\tilde{R}_{\Pi_{n}}])$ where $\tilde{R}_{\Pi_i}$ denotes the ${\Pi_i}$-th diagonal block in $\tilde{R}_x(t+1)$.

Giving the new targets $x^c=[x^c_1,...,x^c_k]$ at instant $t+1$, we can write the updated state vector $s'(t+1)= \left[\begin{array}{ccc}\overbrace{x^c_1,...,x^c_k}^{x^c} & \overbrace{x_{\pi_{d+1}},...,x_{\pi_n},a}^{s^r} \end{array}\right]$.  Assume the new targets are distributed as $x^c \sim [R^c, z^c]$ where $R^c$ denotes a block-diagonal \emph{noninformative} information matrix\footnote{$R^c =\mbox{diag}(\epsilon I,...,\epsilon I)$ and $z^c=[0,...,0]^T$ where $\epsilon$ denotes a small positive number.}. In addition we assume $x^c$ be statistically independent of $s^r$. Therefore, the updated state vector $s'(t+1)$ is distributed as
\begin{equation}
    s'(t+1) \sim  \left[\begin{array}{cccc} R^c & 0 & 0 & z^c \\ 0 & \tilde{R}_x^r & \tilde{R}_{xa}^r & \tilde{z}^r \\
     0 & 0 & \tilde{R}_a(t+1) & \tilde{z}_a(t+1)\end{array}\right]
     \label{Eq:ChangeDim}
\end{equation}

\section{AN ILLUSTRATIVE EXAMPLE} \label{SC:5}
\begin{figure}[thbp]
    \centering
    \vspace{-.1in}
    \includegraphics[width=8cm]{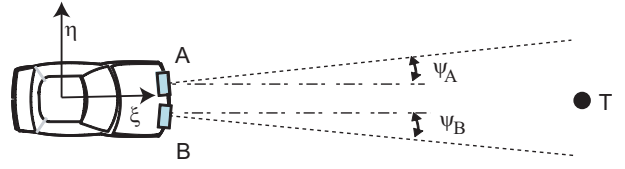}
    \caption{\protect\small  An example}
    \vspace{-.15in}
    \label{FG:example}
\end{figure}

The schematic illustration in Fig.~\ref{FG:example} includes the sensors mounted at the front of a vehicle at positions A and B.  A single target T, in front and in the same lane as the vehicle, moves away from the vehicle.

In the scenario illustrated in Fig.~\ref{FG:example}, the positions of the sensors A and B are denoted by $(\xi_{A0},\eta_{A0})$ and $(\xi_{B0},\eta_{B0})$, respectively. The orientations of the sensors A and Sensor B are denoted by $\Psi_{A0}$ and $\Psi_{B0}$, respectively. The target is located at position $x=\left( {\xi}, v_{\xi}, {\eta}, v_{\eta} \right)$ in the $\xi\eta$-coordinate system. Let the registration parameters of Sensor A and B be $a_A=(\xi_{A0},\eta_{A0}, \Psi_{A0})$ and $a_B=(\xi_{B0},\eta_{B0}, \Psi_{B0})$, respectively. Then the joint state vector $s=\left[\begin{array}{ccc} x &a_A &a_B\end{array}\right]$.

Providing each sensor measures range ($r$), range rate ($\dot{r}$), and azimuth angle ($\theta $), the measurement functions $h_A=( {r_{A} }, {\dot{r}_{A} }, \theta_{A})$ for Sensor A can be expressed as, $r_{A} =\sqrt{(\xi-\xi_{A0} )^{2} +(\eta-\eta_{A0} )^{2} } $, $\dot{r}_{A} =v_{r} ^{T} n_A$ and $\theta _{A} =\arctan (\frac{\eta-\eta_{A0} }{\xi-\xi_{A0} } )-\Psi _{A0} $, respectively, where $v_{r}$ and $n_A$ denote the relative velocity vector $(v_\xi, v_\eta)$ and unit vector along the direction from the target to Sensor A.  Approximating the measurement functions $h_A$ in linear form at the neighborhood of the point $(x^{*},a_A^*)$, produces:
{\scriptsize
\[ \label{GrindEQ__2_24_)}
C_{Ax} =\left. \frac{\partial h_A}{\partial x } \right|^{*} =\left[\begin{array}{cccc} {\frac{\xi^{*} -\xi_{A0}^* }{r_A^{*} } } & {0} & {\frac{\eta^{*} -\eta_{A0}^* }{r_A^{*} } } & {0} \\ {0} & {\frac{\xi^{*} -\xi_{A0}^* }{r_A^{*} } } & {0} & {\frac{\eta^{*} -\eta_{A0}^* }{r_A^{*} } } \\ {-\frac{\eta^{*} -\eta_{A0}^* }{(r_A^{*} )^{2} } } & {0} & {\frac{\xi^{*} -\xi_{A0}^* }{(r_A^{*} )^{2} } } & {0} \end{array}\right]
\]
\[ \label{GrindEQ__2_25_)}
c_{Aa} =\left. \frac{\partial h}{\partial a_{A} } \right|^{*} =\left[\begin{array}{ccc} -{\frac{\xi^{*} -\xi_{A0}^* }{r_A^{*} } }& -{\frac{\eta^{*} -\eta_{A0}^* }{r_A^{*} } } & {0}\\ {0} & 0& 0\\ 0 &0 & {-1}\end{array}\right]
\]}
where $x^{*}=(\xi^*,v_\xi^*,\eta^*,v_\eta^*)$, $a_A^*=(\xi_{A0}^*,\eta_{A0}^*,\Psi_{A0}^*)$, and $r_{A}^* =\sqrt{(\xi^*-\xi_{A0}^* )^{2} +(\eta^*-\eta_{A0}^* )^{2}}$.

Similarly, $h_B$, $C_{Bx} $, and $c_{Ba} $ of the sensor B can be derived.

Let the measurement vector \textit{o} be denoted as
\begin{equation} \label{GrindEQ__2_27_}
o=(\overbrace{ {r_{A} }, {\dot{r}_{A} }, {\theta _{A} }}^{o_A},  \overbrace{{r_{B} }, {\dot{r}_{B} }, {\theta _{B}}}^{o_B})^{T}
\end{equation}
The matrix coefficients $C_{x}$ and $C_{a}$ of the measurement equation in  \eqref{ZEqnNum185990} for the scenario illustrated in Fig.~\ref{FG:example} can then be expressed as:

\begin{equation} \label{GrindEQ__2_28_}
C_{x} =\left[\begin{array}{c} {C_{Ax} } \\ {C_{Bx} } \end{array}\right]
\end{equation}

\begin{equation} \label{GrindEQ__2_29_}
C_{a} =\left[\begin{array}{cc} {c_{Aa} } & {0} \\ {0} & {c_{Ba} } \end{array}\right]
\end{equation}

Assuming the system dynamics can be modeled with a constant velocity (CV) model,   $\Phi _{x}$, $G_{x}$, $u_2$ in  \eqref{GrindEQ__2_14_} can be written as
\begin{equation} \label{GrindEQ__2_30_}
\Phi _{x} =\left[\begin{array}{cccc} {1} & {\Delta T} & {0} & {0} \\ {0} & {1} & {0} & {0} \\ {0} & {0} & {1} & {\Delta T} \\ {0} & {0} & {0} & {1} \end{array}\right]
\end{equation}
\begin{eqnarray} \label{GrindEQ__2_31_}
G_{x} =I_4 & & u_2 = 0
\end{eqnarray}

Process noise $w$ in \eqref{GrindEQ__2_14_} is denoted by the information array $[R_{w} ,z_{w}]$, expressed as $z_{w}=0$ and
\begin{equation}
    R_{w} = \left[\begin{array}{cc} q_\xi W & 0\\0 & q_\eta W\end{array}\right] \label{EQ:proces_noise}
\end{equation}
where $W = \left[\begin{array}{cc} \sqrt{\frac{\Delta T^3}{3}} & \sqrt{3\frac{\Delta T}{4}}\\  0 & \sqrt{\frac{\Delta T}{4}} \end{array}\right]$; $q_\xi$ and $q_\eta$ are random walking parameters for $\xi$-coordinate and $\eta$-coordinate, respectively.

\section{Simulation} \label{SC:6}
As shown in Fig.~\ref{FG:example}, the simulation results presented here are based on two sensors simulated near the front bumper. Sensor A is located at $(\xi_{A0}=2,\eta_{A0}=0.6)$, oriented $10^\circ$ ($\Psi_{A0}=10^\circ$) outwards from the vehicle's bore-sight ($\xi$-coordinate). Sensor B is located at $(\xi_{B0}=2,\eta_{B0}=-0.6)$, oriented $10^\circ$ ($\Psi_{B0} = -10^\circ$) outwards from the vehicle's bore-sight. The random walking parameters in \eqref{EQ:proces_noise} are expressed as $q_\xi=0.1$ and  $q_\eta=0.1$. The measurement noise variance $(\sigma_r, \sigma_{rr}, \sigma_{\theta})$ is set to $(0.1, 0.2, 1^\circ)$. Three algorithms are implemented: SEP, UKF, and FMAP. SEP \cite{Sviestins2002} treats the tracking and registration problems separately; but the UKF and FMAP algorithms address the problems jointly. Each algorithm has been implemented in Matlab on a 2 GHz Intel Core 2 Duo processor running Window XP. No special care has been taken to produce efficient code.

\begin{figure}[thbp]
    \includegraphics[width=8.5cm]{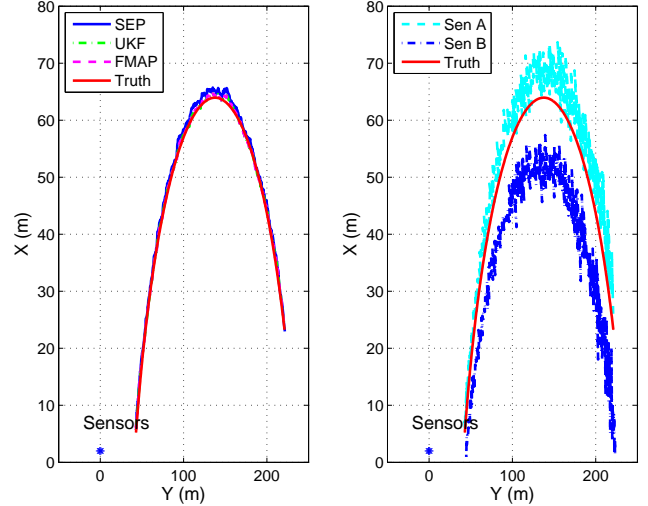}
    \caption{\protect\small  Top-down view of the trajectories of registered (left) and unregistered (right) sensor measurements of an tracked target vs. ground truth, respectively}
    \label{FG:topdown}
\end{figure}

\begin{figure}[htb]
  \centering
\includegraphics[width=0.46\textwidth]{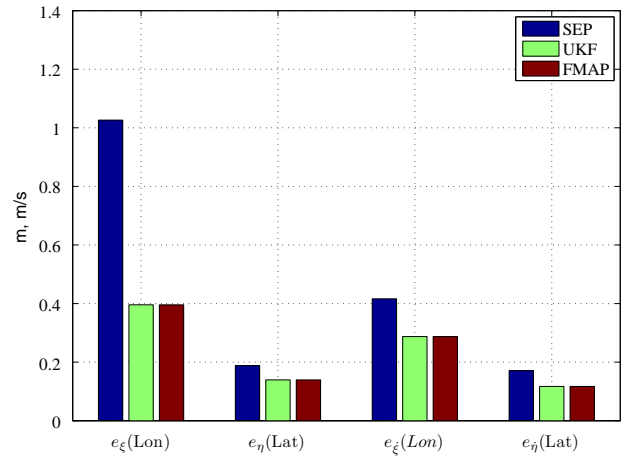}
\caption{\protect\small Mean errors of estimated longitudinal and lateral positions ($e_\xi$ and $e_\eta$) and the corresponding velocities ($e_{\dot\xi}, e_{\dot\eta}$) for a tracked target.}
\label{FG:track_error_sigma}
\end{figure}

\begin{figure}[htb]
    \centering
    \includegraphics[width=7.4cm]{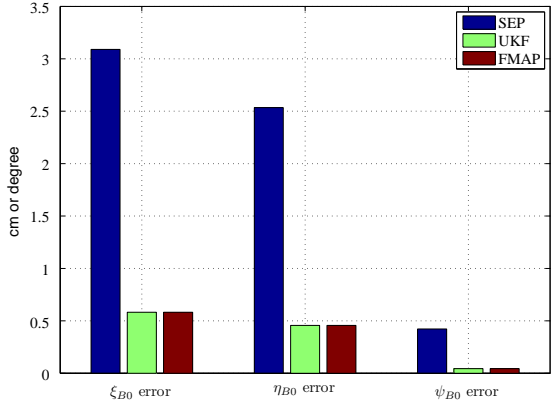}
    \caption{\protect\small  Mean errors of sensor registration estimates for Sensor B.}
    \label{FG:alignment_sigma}
\end{figure}

In the first simulation, ten targets are randomly generated in the field-of-view of the sensors over a period of about 50 seconds. We initially set the registration parameters of the sensors randomly with small numbers. Tracks are initialized using the zero-mean noninformative distributions. Fig.~\ref{FG:topdown} shows the trajectories of unregistered measurements and an tracked target vs. ground truth. The red solid line represents the true trajectory of the target while the other blue solid, green dash-dotted, and dashed lines represent the fused results of the SEP, UKF, and FMAP algorithms, respectively. Fig.~\ref{FG:track_error_sigma} show the mean errors\footnote{The mean error is defined as $e_x = E\{|x - x_T|\}$ with $x$ and $x_T$ the measurement and ground truth, respectively. Giving a series of measurement and truth pairs $(x_i, x_{Ti})$, $1\leq i\leq N$, the sample mean error is computed as $e_x = \frac{\sum_{i}^N | x_i - x_{Ti}|}{N}$.} of position and velocity estimates of the target for the three algorithms, respectively. The bar plot in Fig.~\ref{FG:alignment_sigma} shows the mean errors of the registration estimates for the three algorithms. The performance difference between UKF and FMAP are too small to seen on the scale used in the Figures. Note that in all the cases the performance of FMAP and UKF is clearly superior to that of SEP. This confirms the asymptotical optimality of JTR algorithm versus the decoupled approaches (e.g., SEP algorithm).

\begin{figure}[htb]
    \centering
    \includegraphics[width=8.4cm]{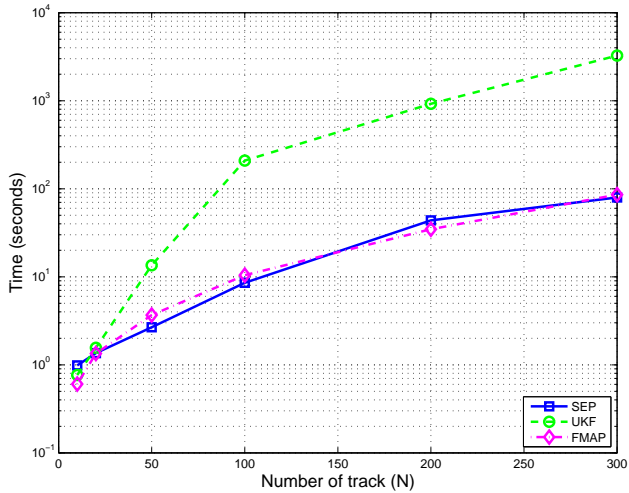}
    \caption{\protect\small  Execution time}
    \label{FG:complexity}
\end{figure}

In the next simulation, we investigate the time complexity of the SEP, UKF, and FMAP algorithms by varying the number of simulated targets from 10 to 300. Fig.~\ref{FG:complexity} plots the execution time curves for the three algorithms at the different input sizes. The figure clearly illustrates the factor that the time complexities of FMAP and SEP are similar and are both an order of magnitude lower than that of UKF. This confirms the superior time-efficiency of FMAP vis-a-vis that of UKF.

\begin{figure}[htb]
    \centering
    \includegraphics[width=8.4cm]{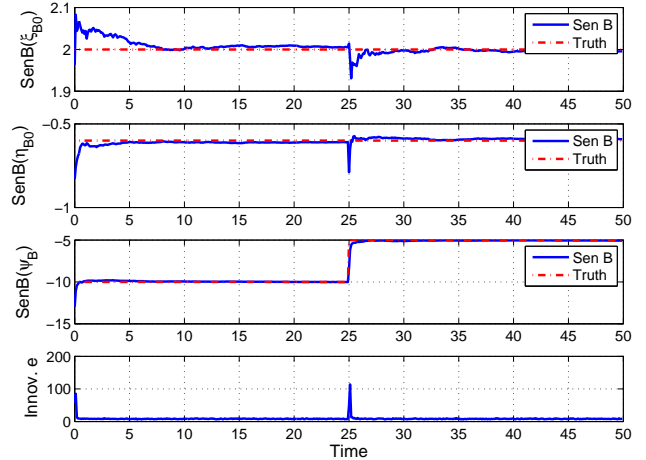}
    \caption{\protect\small  The response under step change of sensor registration. The top three plots are for the estimates of the registration parameters $\xi_{B0}$, $\eta_{B0}$, and $\Psi_{B0}$ vs. ground truth. The bottom plot is the innovation error. }
    \label{FG:dynamicregistr}
\end{figure}

The last simulation investigates how FMAP responses in the step change of registration. For example, in Fig.~\ref{FG:dynamicregistr} the orientation of Sensor B ($\Psi_{B0}$) steps from $-10^\circ$ to $-5^\circ$ at Second 25 (see the third plots from the top), namely, the sensor rotates counterclockwise $5^\circ$ that may be caused by an external collision impact. As in the first simulation, ten targets are randomly generated for the 50-second simulation. The innovation error curve has two peaks caused by two different cases: unknown registration at Second 0 and step change of registration at Second 25, respectively. As shown in Fig.~\ref{FG:dynamicregistr}, the registration estimates converge to their corresponding ground truth within 5 seconds for both cases.

\section{Vehicular Experiment} \label{SC:Experiment}
\begin{figure}[h]
 \centerline{
    \includegraphics[width=4.2cm,height=2.6cm]{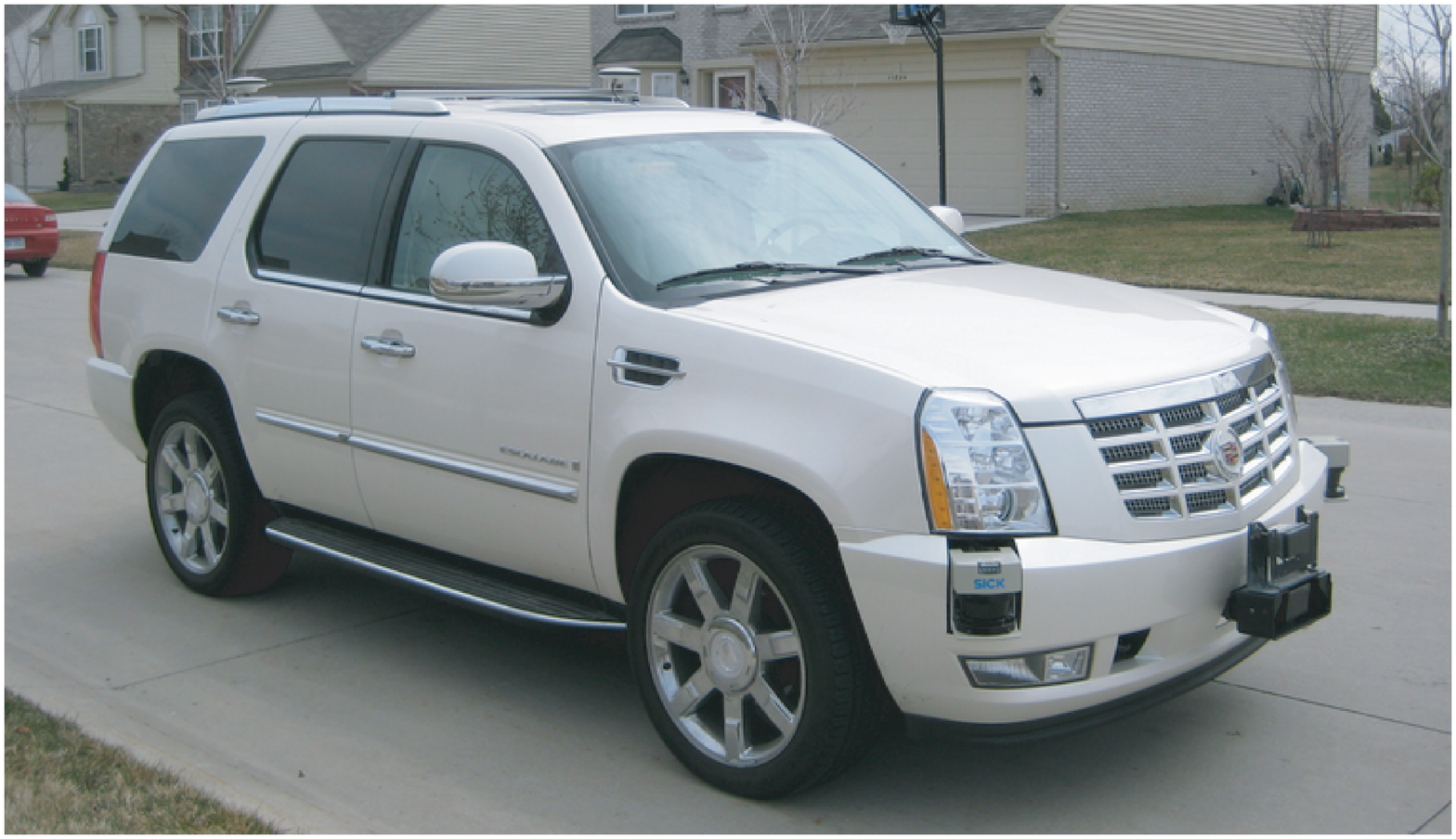}
    \includegraphics[width=4.2cm,height=2.6cm]{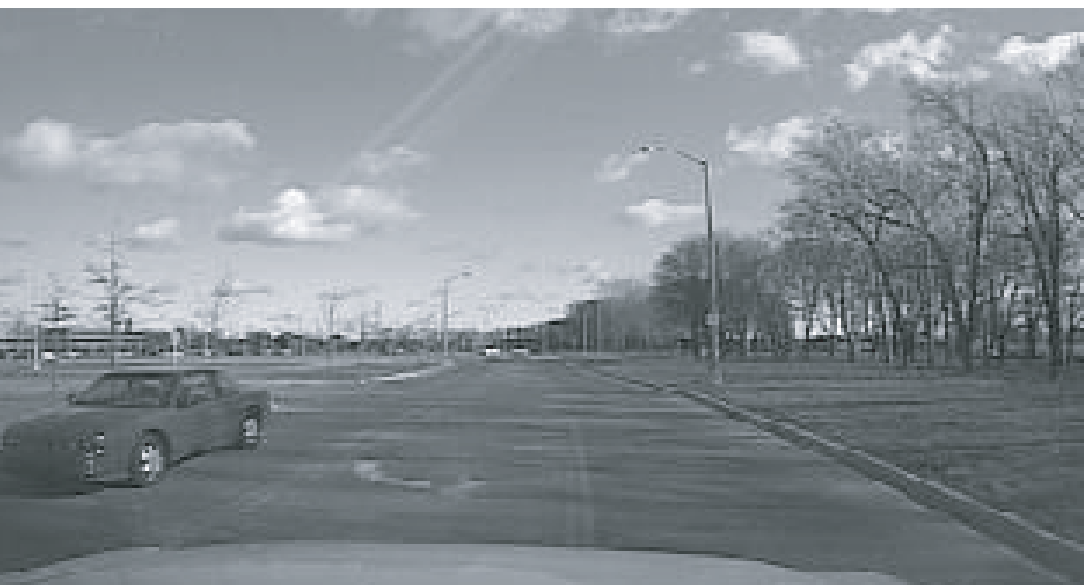}}
 \centerline{
    \makebox[4.2cm] {(a)} \makebox[4.2cm]{(b)}
 }
 \centerline{
    \includegraphics[width=7.0cm]{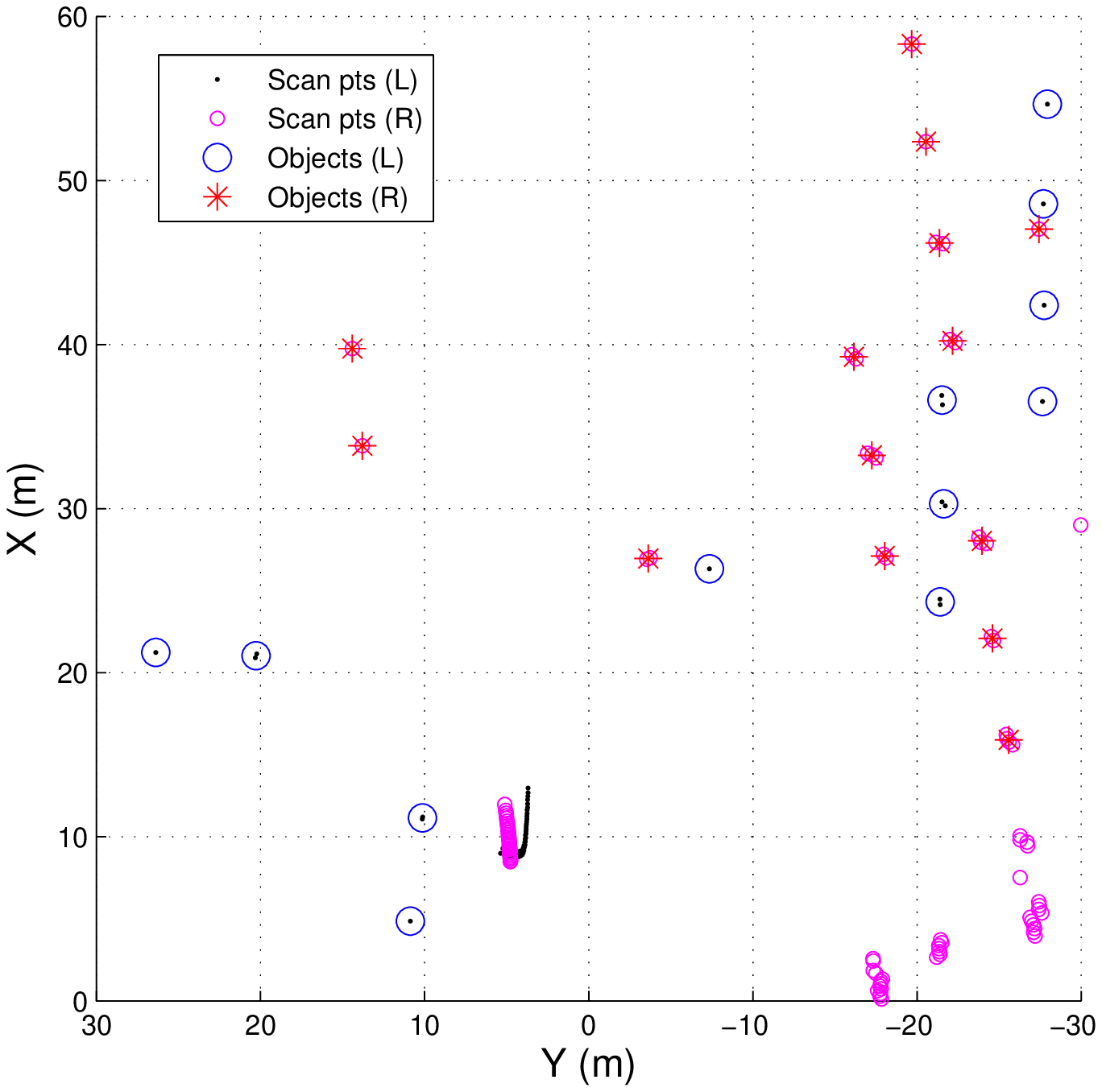}
    }
 \centerline{
    \makebox[8.4cm] {(c)}
 }
  \centerline{
    \includegraphics[width=7.0cm]{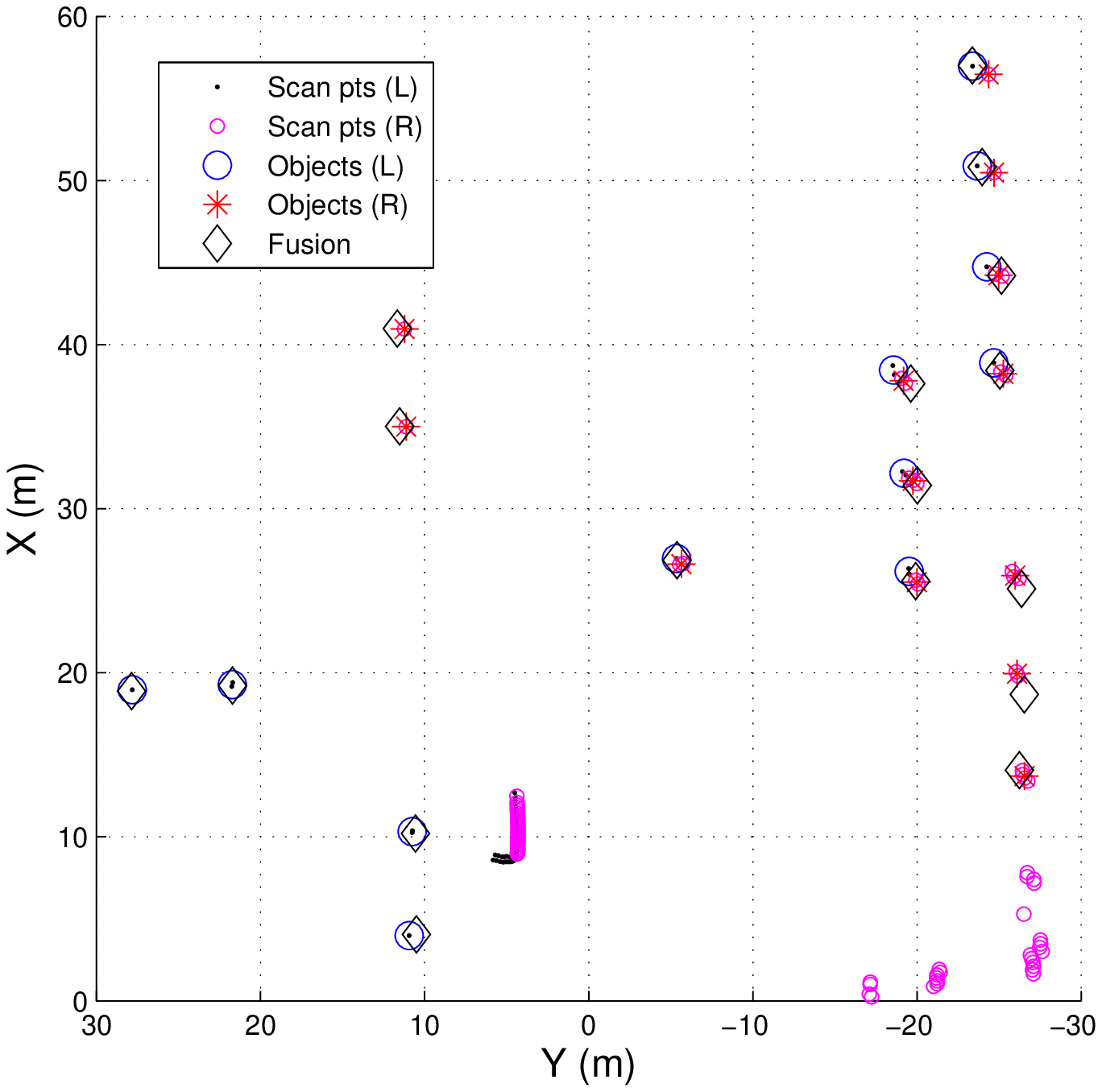}
    }
 \centerline{
    \makebox[8.4cm] {(d)}
 }
\caption{\protect\small  (a) The test-bed vehicle equipped with two rangefinders. (b) The snapshot of a scene in the parking lot. (c) The top-down view of unregistered range data of the scene in (b). (d) The top-down view of registered range data of the scene in (b).}
    \label{FG:testbed}
\end{figure}

A test vehicle (Fig. \ref{FG:testbed} (a)) equipped with two SICK LMS291 rangefinders is used as the test-bed to verify the effectiveness of the proposed FMAP. The two rangefinders are mounted at two corners of the front bumper. The positions and orientations of the rangefinders are precisely surveyed. The left rangefinder (Sensor A) is located at $(\xi_{A0}=3.724,\eta_{A0}=0.883)$, oriented $45^\circ$ ($\Psi_{A0}=45^\circ$) outwards from the vehicle's bore-sight ($\xi$-coordinate). The right rangefinder (Sensor B) is located at $(\xi_{B0}=3.720,\eta_{B0}=-0.874)$, oriented $45^\circ$ ($\Psi_{B0} = -45^\circ$) outwards from the vehicle's bore-sight. The vehicle is also equipped with wheel encoders and an IMU sensor for determining the motion control inputs.

The rangefinder scans from right to left in its $180^\circ$ field-of-view at a resolution of $0.5^\circ$ and generates 361 distance measurements to the closest line-of-sight obstacles. Each rangefinder is equipped with a point-object detector (e.g., tree trunks, light poles, and etc.) that selects scan clusters such that 1) the size of the cluster is less than 0.5 meter, and 2) the distance to the nearest point of other clusters is larger than 5 meters. The speed (i.e., range rate) of a point object is inferred from the host vehicle motion because of the stationary assumption of the point objects.

The experiment~\footnote{The demo source code can be downloaded from \url{http://code.google.com/p/joint-calibration-tracking/} along with the data set.} was conducted in a parking lot with plenty road-side point objects shown in Fig. \ref{FG:testbed} (b). The vehicle was manually driven, and an observation sequence of about 50 seconds was collected. The plot (c) shows a snapshot of unregistered range data for the scene in Fig. \ref{FG:testbed} (b). The black dots and magenta circles in the plot denote the scan data generated by the left and right rangefinders, respectively. The blue circles and red stars denote the detected point objects from the left and right rangefinders, respectively.

The measurement inconsistence caused by the registration bias is clearly seen. For example, the contours of the vehicle and the point objects in (b) from the sensors is not aligned. The inconsistence raises an issue for data association. Without a consistent registration, data from one sensor cannot be correctly associated with that from a different sensor.
For comparison, the snapshot of the registered range data of the same scene using FMAP algorithm is shown in Fig. \ref{FG:testbed} (d). The additional black diamonds in (d) denote the fused targets. This plot clearly reveals the fact that FMAP significantly improves the sensor registration and thus the data association among multiple sensors.

\begin{figure}[htb]
    \centering
 \subfigure[]{\includegraphics[width=8.6cm]{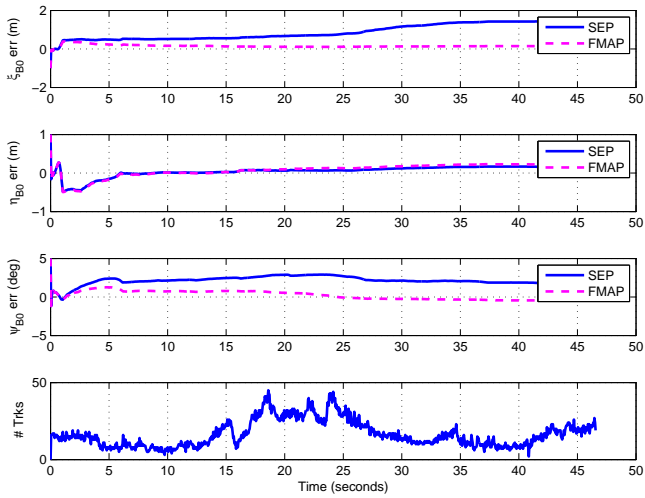}}
 \subfigure[]{\includegraphics[width=7.6cm]{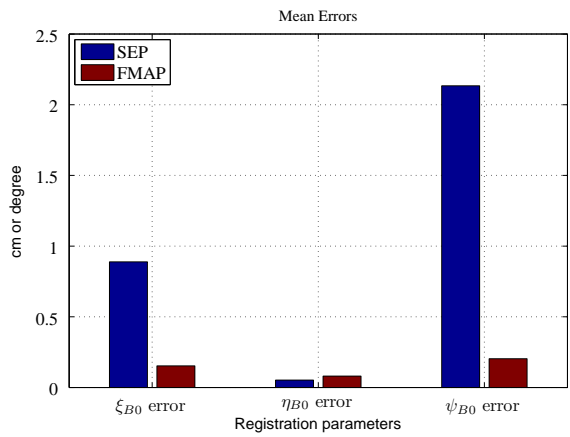}}
    \caption{\protect\small (a) The error curves of sensor registration estimates ($\xi_{B0}$, $\eta_{B0}$ and $\Psi_{B0}$) for the right rangefinder when the true value is initially unknown. The last plot shows the number of detected objects varies with time. (b) The corresponding mean errors for the registration estimates.}
    \label{FG:vehicle_1}
\end{figure}

In each discrete time instant, the detected objects from both sensors are matched with the tracked targets (drawn as black diamonds in Fig. \ref{FG:testbed} (d)) at previous instant using a nearest-neighbor and a maximum distance gating data association strategy. We then apply the FMAP filter to jointly track the associated track-object pairs and registration parameters. Note that no target is visible in the whole observation sequence. As shown in the last plot in Fig. \ref{FG:vehicle_1}, the dimension of the state vector $s$ changes due to new targets coming into or existing targets leaving the field-of-view of the sensors. The method described in Section \ref{SC:change_dim} is used to handle the change.

In the experiment, the right rangefinder's registration is unknown and initially set to $(\xi_{B0}=2.720,\eta_{B0}=-0.126,\Psi_{B0} = -40^\circ)$. The SEP and FMAP algorithms are applied to the data set. The resulted registration error curves for both algorithms versus the surveyed values are shown in Fig.~\ref{FG:vehicle_1}(a). The blue solid lines and the magenta dashed lines denote the error curves of SEP and FMAP, respectively. FMAP clearly exhibits superior performance comparing with that of SEP. A shown in Fig. \ref{FG:vehicle_1}(a), the estimates of the sensor's position $(\xi_{B0},\eta_{B0})$ and orientation $\Psi_{B0}$ converge to their true values in about 5 seconds, respectively. The slight transitory oscillation near the start of the run is partially due to the vehicle's steering maneuver and lack of shared detected objects from both sensors. Selecting a suitable model for the target motion requires giving consideration to the tradeoff between the filter accuracy and the model complexity. Better results can be obtained by incorporating vehicular motion model and global target map building. This may increase the computational complexity of the algorithm. The fourth plot in \ref{FG:vehicle_1}(a) shows the number targets tracked by FMAP.

On the other hand, the significant mean registration error by the SEP algorithm can be observed in Fig.~\ref{FG:vehicle_1}(b). We should note that the divergence of SEP curves increases over time in Fig.~\ref{FG:vehicle_1}(a), and this is partially contributed by the erroneous data association caused by registration biases of previous cycles. Therefore, the estimates of registration by FMAP is more consistent and robust than that by SEP.

\section{SUMMARY AND CONCLUSION} \label{SC:7}
In this article, we have addressed the recursive JTR. The FMAP algorithm is derived in which the time complexity scales linearly with the numbers of measurements and targets. It is proved that FMAP is asymptotically optimal and has an $O(n)$ implementation based on matrix orthogonal factorization. The results from experiments on synthetic and vehicular data demonstrate that, as expected, FMAP consistently performs better than the methods where tracking and registration are separately treated. It has been also demonstrated experimentally using the synthetic data that the complexity of FMAP is indeed $O(n)$. Although FMAP and UKF are quite equivalent in performance but the time complexity of FMAP is a magnitude lower than that of UKF. Additionally, results of simulation and vehicle experiment demonstrate that FMAP can handle the dynamics of sensor registration and a variable number of targets.

\bibliographystyle{IEEEtranS}
\bibliography{gm_eci}

\section*{Appendix A}
\begin{lemma} \label{lemma_Naik}
Let $\rho \sim \mathcal{N}(\mu_\rho, \Sigma_\rho)$ or in information array terms, $[R_\rho, z_\rho]$. If $\omega=\alpha \rho + \beta$, $\omega\sim \mathcal{N}(\alpha \mu_\rho + \beta, \alpha \Sigma_\rho \alpha^t)$, or in information array terms, $\omega$ can be represented through $[R_\omega, z_\omega]$, where $R_\omega = R_\rho\alpha^{-1}$ and $z_\omega = z_\rho + R_\omega\beta$.
\end{lemma}
\begin{proof}
Since $\mu_\rho = R_\rho^{-1}z_\rho$ and $\Sigma_\rho=R_\rho^{-1}R_\rho^{-t}$, we get
\begin{eqnarray}
\mu_\omega &=& R_\omega^{-1}z_\omega=\alpha \mu_\rho + \beta\nonumber
\end{eqnarray}
and
\begin{eqnarray}
\Sigma_\omega &=& R^{-1}_\omega R^{-t}_\omega = \alpha \Sigma_\rho \alpha^t = \alpha R^{-1}_\rho R^{-t}_\rho \alpha^{t} \nonumber
\end{eqnarray}
One can verify that $R_\omega = R_\rho \alpha^{-1} $ and
\begin{eqnarray}
z_\omega &=& R_\omega \mu_\omega = R_\omega (\alpha \mu_\rho + \beta) = R_\omega (\alpha R_\rho^{-1} z_\rho + \beta)\nonumber\\
&=& R_\rho \alpha^{-1} (\alpha R^{-1}_\rho z_\rho + \beta)\nonumber\\
&=& (R_\rho \alpha^{-1})(\alpha R^{-1}_\rho) z_\rho + (R_\rho \alpha^{-1})\beta\nonumber\\
&=& z_\rho + (R_\rho \alpha^{-1})\beta\nonumber\\
&=& z_\rho + R_\omega \beta \nonumber
\end{eqnarray}
\end{proof}

\begin{lemma} \label{lemma_NAIK2}
 Let the system dynamics be defined in \eqref{GrindEQ__2_14_}; the density function of the state variable $s(t)=[x,a]$ be expressed as information array in \eqref{ZEqnNum934500}; and random noise term $w$ be in information array form $[R_w, z_w]$. Let $\rho \equiv [w, s(t)]^t$. If $w$ and $s(t)$ are statistically independent, the joint density of $\omega \equiv [w, x(t+1), a(t+1)]^t$ can be represented through the information array $[R_\omega, z_\omega]$, where
\[
R_\omega=\left[\begin{array}{ccc} {R_{w}} & {0} & {0} \\ {-\hat{R}_{x} \Phi _{x}^{-1} G_{x} } & {\hat{R}_{x} \Phi _{x}^{-1} } & {\hat{R}_{xa}} \\ {0} & {0} & {\hat{R}_{a} } \end{array}\right]
\]
and \[
z_\omega=\left[\begin{array}{c} {z_{w} } \\ {\hat{z}_{x} +\hat{R}_{x} \Phi _{x}^{-1} u_2} \\ {\hat{z}_{a}} \end{array}\right]
\]
\end{lemma}
\begin{proof}
Since $w$ and $s(t)$ are statistically independent, the information array of the joint vector $\rho = [w, s(t)]^t=[w, x, a]^t$ is represented as $[R_\rho, z_\rho]$ where
$
R_\rho = \left[\begin{array}{ccc}{R_w} & 0 & 0  \\ {0} & {\hat{R}_{x} } & {\hat{R}_{xa} } \\ {0} & {0} & {\hat{R}_{a} }  \end{array}\right]
$ and $
z_\rho = \left[\begin{array}{ccc}  {z_w} & {\hat{z}_{x} } & {\hat{z}_{a} } \end{array}\right]^t$

One verifies that the system dynamics equation \eqref{GrindEQ__2_14_} can be reorganized as
\[
\omega = \alpha \rho + \beta
\]
with $\alpha=\left[\begin{array}{ccc} I & 0 & 0 \\ G_x & \Phi_x & 0 \\ 0 & 0 & I\end{array}\right]$ and $\beta = \left[\begin{array}{c} 0 \\ u_2 \\ 0 \end{array} \right]$,
and $\alpha^{-1} = \left[\begin{array}{ccc} I & 0 & 0 \\ -\Phi_x^{-1}G_x & \Phi_x^{-1} & 0 \\ 0 & 0 & I\end{array}\right]$.

Using Lemma \ref{lemma_Naik}, we obtain the
\[
    R_\omega =  R_\rho \alpha^{-1} = \left[\begin{array}{ccc} {R_{w}} & {0} & {0} \\ {-\hat{R}_{x} \Phi _{x}^{-1} G_{x} } & {\hat{R}_{x} \Phi _{x}^{-1} } & {\hat{R}_{xa}} \\ {0} & {0} & {\hat{R}_{a} } \end{array}\right]
\]
and
\[
z_\omega = z_\rho + (R_\rho \alpha^{-1})\beta = \left[\begin{array}{c} {z_{w} } \\ {\hat{z}_{x} +\hat{R}_{x} \Phi _{x}^{-1} u_2} \\ {\hat{z}_{a}} \end{array}\right]
\]
\end{proof}

\begin{lemma} \label{lemma_marginal}
Let $\rho=\left[\begin{array}{c}\rho_1\\\rho_2\end{array}\right]$ be distributed as $\rho \sim \mathcal{N}(\mu_\rho, \Sigma_\rho)$ or information array $\left[R_\rho,z_\rho \right]$ with $R_\rho=\left[\begin{array}{cc}R_{\rho_{11}} & R_{\rho_{12}}\\0 & R_{\rho_{22}}\end{array}\right]$ and $z=\left[\begin{array}{c}z_{\rho_1}\\z_{\rho_2}\end{array}\right]$. Then the marginal distribution of $\rho_2$ is normal and has $\rho_2 \sim \left[R_{\rho_{22}}, z_{\rho_2}\right]$.
\end{lemma}
\begin{proof}
$R_\rho^{-1} = \left[\begin{array}{cc}R_{\rho_{11}}^{-1} & -R_{\rho_{11}}^{-1}R_{\rho_{12}}R_{\rho_{22}}^{-1} \\0 & R_{\rho_{22}}^{-1}\end{array}\right]$. Thus the mean \begin{equation}\begin{split}
\mu_\rho &= R_\rho^{-1} z_\rho\\
& = \left[\begin{array}{cc}R_{\rho_{11}}^{-1} & -R_{\rho_{11}}^{-1}R_{\rho_{12}}R_{\rho_{22}}^{-1} \\0 & R_{\rho_{22}}^{-1}\end{array}\right]\left[\begin{array}{c}z_1\\z_2\end{array}\right]\\
& = \left[\begin{array}{c} R_{\rho_{11}}^{-1} z_{\rho_1}  -R_{\rho_{11}}^{-1}R_{\rho_{12}}R_{\rho_{22}}^{-1}z_2 \\  R_{\rho_{22}}^{-1}z_{\rho_2} \end{array}\right]
\end{split}
\nonumber
\end{equation}
The marginal distribution of $\rho_2 \sim \mathcal{N}(\mu_{\rho_2}, \Sigma_{\rho_2})$ with $\mu_{\rho_2} = R_{\rho_{22}}^{-1}z_{\rho_2}$ and $\Sigma_{\rho_2} = R_{\rho_{22}}^{-1}R_{\rho_{22}}^{-t}$. Therefore $\rho_2 \sim \left[R_{\rho_{22}}, z_{\rho_2}\right]$.
\end{proof}

%
%

\balance

\end{document}